\documentclass[10pt,twocolumn,letterpaper]{article}

\usepackage[pagenumbers]{wacv} 

\newcommand{\method}{\textbf{Open-V}}

\newif\ifarxiv
\arxivfalse
\usepackage{float}
\definecolor{wacvblue}{rgb}{0.21,0.49,0.74}
\usepackage[pagebackref,breaklinks,colorlinks,allcolors=wacvblue]{hyperref}

\usepackage{listings}
\def\wacvPaperID{*****} 
\def\confName{WACV}
\def\confYear{2026}

\title{Training-Free Generalized Few-Shot Segmentation through Open-Vocabulary Semantic Arbitration}

\author{Silas Kwabla Gah\\
Department of Computer Science\\
University of Ghana\\
{\tt\small skgah001@st.ug.edu.gh}
\and
Ebenezer Owusu\\
Department of Computer Science\\
University of Ghana\\
{\tt\small ebeowusu@ug.edu.gh}
}

\begin{document}
\maketitle
\begin{abstract}
Generalized Few-Shot Semantic Segmentation (GFSS) has traditionally been approached as a 
representation-learning problem, requiring task-specific adaptation to incorporate novel classes 
from limited support examples. Recent foundation models, however, 
already exhibit strong open-vocabulary recognition and segmentation capabilities, raising 
a different question: can GFSS be solved through inference-time coordination of frozen 
semantic priors rather than parameter adaptation? We answer this question with Open-V, 
a training-free GFSS framework that combines Segment Anything (SAM3) Promptable Concept 
Segmentation (PCS) with a $K$-shot CLIP support centroid through 
calibrated per-pixel semantic arbitration. Open-V introduces no trainable components 
and supports arbitrary semantic categories at inference time.
Beyond segmentation performance, our study contributes three broader findings. First, 
we show that support information can be incorporated through inference-time semantic 
grounding, and that its contribution increases as foundation-model text priors weaken 
on label-disjoint vocabularies. Second, we identify a reproducibility confound in
foundation-model segmentation, demonstrating that preprocessing and evaluation-space 
mismatches can silently distort reported performance. Finally, we validate Open-V across 
PASCAL-5i, COCO-20i, and ADE-OW, showing that training-free coordination of foundation-model 
priors generalizes across both conventional GFSS and open-vocabulary evaluation settings.
On PASCAL-5$_i$ (1-shot), Open-V attains base/novel/harmonic
mIoU of 78.4/77.5/77.9, without GFSS-specific training surpassing the strongest trained baseline by
$+$17.7 HM.
\end{abstract}

\section{Introduction}
\label{sec:intro}
 
Semantic segmentation, the task of assigning a class label to every pixel, has advanced
rapidly through large supervised datasets and deep architectures~\cite{chen2018deeplab}.
Yet two structural limitations persist. First, annotating pixel-level masks
is expensive and slow. Second, models trained on a fixed vocabulary fail on
any class outside that set, no matter how closely related it is to a seen
class.
 
\textbf{Generalized Few-Shot Semantic Segmentation (GFSS)} addresses both
limitations simultaneously. Given $C_b$ base classes with dense annotations
and $C_n$ novel classes described only by $K$ labelled support images, a
GFSS model must segment all $C_b + C_n$ classes simultaneously in a single
forward pass. The joint requirement distinguishes GFSS from vanilla
few-shot segmentation (FSS) approaches~\cite{wang2023focus},
which handle one novel class in isolation and ignore base-class inference entirely.
 
Existing GFSS methods~\cite{tian2022generalized,hossain2024visual,geng2024enhancing}
attack this problem by \emph{parameter adaptation}: a base-class decoder is
trained on abundant base data, then a small visual-prompting head is fitted
for novel classes. This design has two inherent costs. The class vocabulary
is fixed at training time, that is any change to the class list forces
retraining and the novel-class head must generalise from $K$ shots, which
is difficult when the head itself has trainable parameters.
 
Recent visual foundation models offer a different perspective to this task, 
segment anything (SAM) ~\cite{kirillov2023segment} and SAM2~\cite{ravi2024sam2} provide class-agnostic 
mask proposals through geometric prompts. CLIP~\cite{radford2021learning} provides open-vocabulary
recognition through a shared visual-language embedding space.
TFM$^2$~\cite{zhuo2025tfm2} shows that a key-value mask cache built from
few-shot masks can enhance OVSS models without any training, while
LPOSS~\cite{stojnic2025lposs} demonstrates that label propagation over
frozen VLM and vision-model features achieves state-of-the-art
training-free open-vocabulary segmentation. SAM3~\cite{carion2026sam3}
closes a key gap by adding native \emph{Promptable Concept Segmentation
(PCS)}: given a free-form text concept, it returns all instance masks of
that concept with calibrated presence scores, connecting the geometric
precision of SAM with the semantic scope of CLIP.
 
\begin{figure}[!ht]
\centering
\includegraphics[width=\columnwidth]{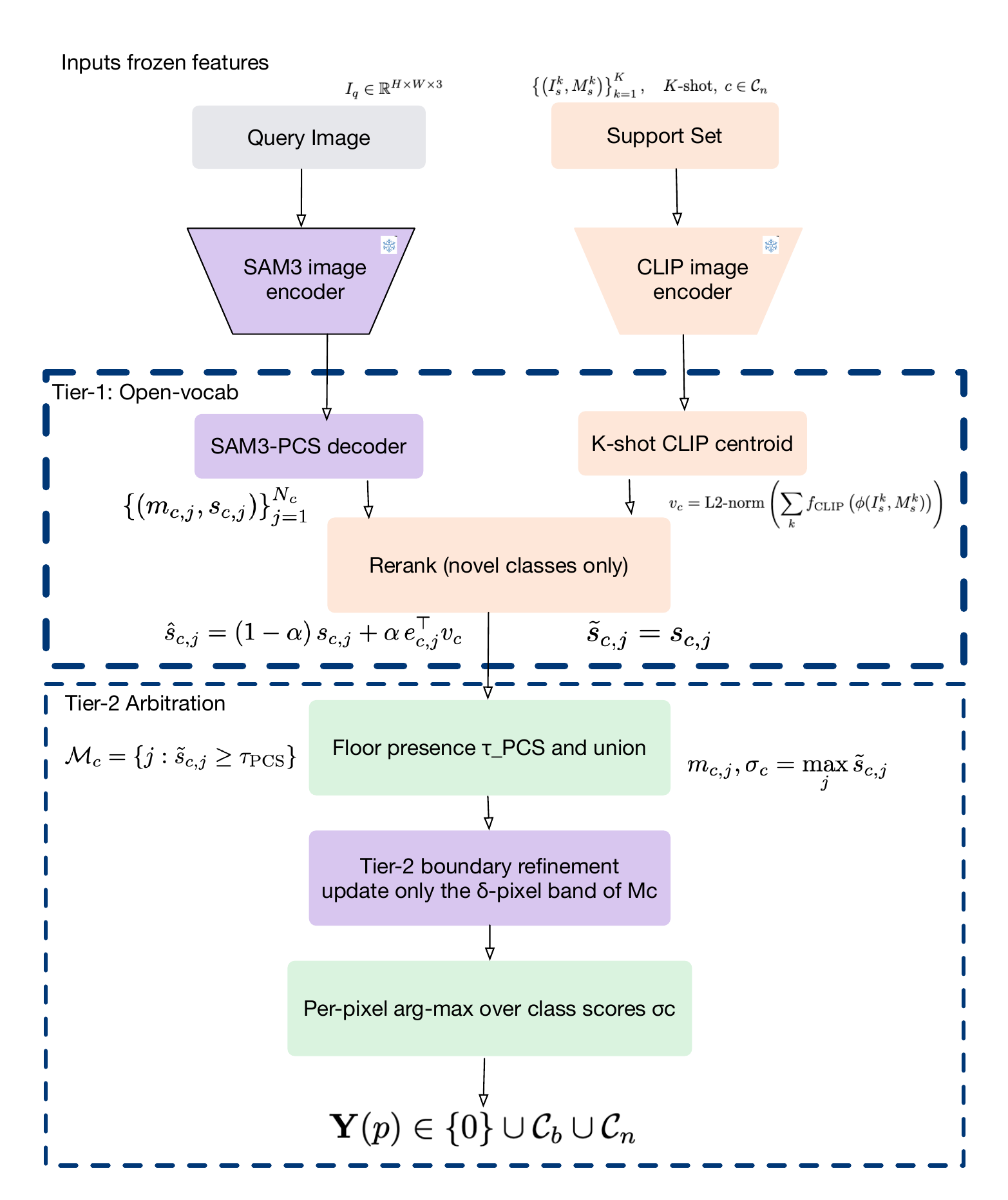}
\caption{\textbf{\method pipeline.} Frozen foundation models (SAM3
ViT-L, CLIP ViT-B/16) carry both evidence sources; the $K$-shot
path on the right is applied only to novel classes; the per-pixel
arg-max at the bottom is the arbitration stage. The query is
encoded once by SAM3; SAM3-PCS decodes one text-conditioned
instance set per class $c\!\in\!\mathcal{C}_b\!\cup\!\mathcal{C}_n$.
Base-class scores pass through unchanged; novel-class scores are
reranked against a single L2-normalised CLIP centroid
$\mathbf{v}_c$ computed once per run. Instances passing the
presence floor $\tau_{\mathrm{PCS}}$ are unioned, sharpened by a
second SAM3 box-prompt call, and resolved by per-pixel arg-max
into the final label map.}
\label{fig:pipeline}
\end{figure}

In this work, we explore a fully frozen SAM3 for GFSS. Instead of introducing new modules
or adapting the model, we argue that PCS makes a conceptual reframing of GFSS not
only possible but natural. Rather than asking ``how do we adapt a
fixed-vocabulary segmenter to absorb new classes?'', we ask: ``how do we
arbitrate, per pixel, among open-vocabulary class hypotheses drawn from
heterogeneous frozen evidence sources?'' This shifts the problem from
parameter fitting to calibrated inference. Three properties follow
immediately: (i) the vocabulary is a property of the arbitration step, not
of the model's weights; (ii) the few-shot signal influences exactly one
evidence source without touching the arbitration rule; and (iii) the
dominant design choices are calibration and combination, not model fitting.
Beyond the proposed framework, we uncover a reproducibility and evaluation confound 
that affects conclusions in foundation-model-based segmentation. 
We show that inconsistencies between preprocessing, prediction, and 
evaluation spaces can silently distort performance estimates, 
obscuring the true impact of support-conditioning mechanisms and segmentation strategies.
By identifying and correcting this confound, we provide a more reliable evaluation protocol 
and highlight the importance of spatial alignment for reproducible foundation-model 
segmentation research. Figure \ref{fig:pipeline} illustrates the Open-V pipeline, 
where frozen SAM3-PCS and CLIP priors are coordinated through 
support-conditioned semantic arbitration and boundary refinement 
to perform training-free generalized few-shot segmentation.

The main contributions of this work are summarized as follows:

\begin{enumerate}[leftmargin=*,topsep=2pt,itemsep=2pt]
\item \textbf{Open-vocabulary semantic arbitration for GFSS.}
We introduce \textbf{Open-V}, a strict training-free GFSS framework that coordinates
frozen SAM3-PCS and CLIP priors through calibrated per-pixel semantic
arbitration. By leveraging foundation-model representations rather than
task-specific adaptation, Open-V naturally supports arbitrary category
vocabularies at inference time.
\item \textbf{Inference-time semantic grounding without parameter updates.}
We show that the few-shot signal can be injected as a post-hoc CLIP
support-centroid rerank rather than a trained prompt head. We further show
that this signal's contribution scales with the degree to which the
foundation text prior weakens: on a label-disjoint vocabulary the optimal
weight of the visual centroid shifts and its peak gain widens by a factor
of 6.6$\times$ relative to in-distribution PASCAL classes.

\item \textbf{Diagnosis of spatial-alignment confounds in foundation-model segmentation.}
We identify and characterize a reproducibility confound that arises when 
foundation-model segmentation pipelines are evaluated under preprocessing and 
coordinate-frame mismatches.
Through controlled analysis across multiple predictor backbones, we show that
inconsistencies between prediction and evaluation spaces can substantially distort
reported performance, highlighting the importance of spatial alignment for reliable
benchmarking of foundation-model-based segmentation systems.
\item \textbf{Cross-benchmark validation across closed- and open-vocabulary regimes.}
We evaluate Open-V on PASCAL-5$_i$, COCO-20$^i$, and ADE-OW, 
demonstrating that training-free coordination of foundation-model priors generalizes 
across conventional GFSS benchmarks and label-disjoint open-vocabulary settings 
without task-specific adaptation.
\end{enumerate}

\section{Related Work}
\label{sec:related}
 
\subsection{Few-Shot and Generalized FSS}
Few-shot semantic segmentation~\cite{tian2020prior,wang2019panet,min2021hypercorrelation,xu2023self,lang2022learning,xu2024eliminating}
conditions on a support set to segment a single novel class per query episode.
Prototypical methods~\cite{tian2020prior,fan2022self,zhang2022feature} compress
support features via mask-average pooling into global prototypes, sacrificing
spatial detail. Pixel-wise matching methods~\cite{min2021hypercorrelation,xu2023self,
peng2023hierarchical}
use cross-attention or 4D convolutions to match support and query features
directly, achieving stronger results at the cost of heavy episodic training.
To eliminate episodic training, recent methods adapt frozen foundation models; for instance, FSS-SAM3~\cite{tsai2026fsssam3} reformulates FSS as a spatial-reasoning task on a shared canvas.

GFSS~\cite{tian2022generalized,hossain2024visual,geng2024enhancing,liu2023cafs}
extends the task to joint base-and-novel inference. BCM~\cite{tian2022generalized}
decomposes the prediction into base and novel streams and uses prototype
matching to handle novel classes. Visual Prompting~\cite{hossain2024visual}
appends learnable visual prompts that adapt to novel classes, achieving the
strongest trained performance on PASCAL-5$_i$. Make It Up~\cite{xie2026make}
augments the support set by synthesising additional shots via diffusion.
All prior GFSS methods fix the vocabulary at training time.
 
\subsection{Foundation Models for Segmentation}
SAM~\cite{kirillov2023segment} demonstrated strong zero-shot class-agnostic
segmentation via point, box, or mask prompts. SAM2~\cite{ravi2024sam2}
extended promptable segmentation to video via a memory-attention mechanism
for temporal tracking. SAM3~\cite{carion2026sam3} introduced Promptable
Concept Segmentation (PCS), enabling segmentation from free-form text
concepts—each returning instance masks with calibrated presence scores—and
natively aligning the geometric precision of SAM with the semantic scope
of vision-language models.
 
CLIP~\cite{radford2021learning} provides a joint visual-language embedding
space, widely used for open-vocabulary classification and retrieval.
DINOv2~\cite{oquab2024dinov2} produces strong dense visual features
for segmentation and retrieval without text supervision.
LPOSS~\cite{stojnic2025lposs} exploits label propagation over DINO and
CLIP features to achieve state-of-the-art training-free open-vocabulary
segmentation, showing that geodesic similarities capture richer
patch-to-patch relationships than Euclidean affinity alone.
 
\subsection{Training-Free Few-Shot and Open-Vocabulary Segmentation}
Unlike traditional closed-set semantic segmentation, 
recent open-vocabulary segmentation methods
~\cite{zhao2017open, xu2022simseg,xu2023san, ding2023open, 
liang2023ovseg,barsellotti2024fossil, barsellotti2024training, 
cha2023learning, cho2024cat, luo2023segclip} 
can recognize arbitrary categories at test time. For instance, 
the pioneering work of Zhao \etal~\cite{zhao2017open} 
learns a joint embedding from visual and word features. 
SimSeg~\cite{xu2022simseg} decouples open-vocabulary segmentation into
 class-agnostic mask generation and mask classification, which SAN~\cite{xu2023san} 
 refines using side-adapter networks.

TFM$^2$~\cite{zhuo2025tfm2} builds a key-value mask cache from a small
number of few-shot masks and shows that three refinement modules—Dynamic
Filter, Channel Reduction, and Feature Alignment—can improve mask
classification in a training-free fashion across multiple OVSS backbones.
VRP-SAM~\cite{sun2024vrpsam} replaces SAM's prompt encoder with a visual
reference adapter trained per dataset. FSSAM~\cite{xu2025fssam} couples
frozen SAM with diffusion-derived features. FSS-SAM3~\cite{tsai2026fsssam3}
reformulates FSS as a spatial-reasoning problem by placing support and
query images on a shared canvas, achieving state-of-the-art 1-shot results
on PASCAL-5$_i$ while revealing that negative prompts consistently degrade
performance in few-shot settings. Matcher~\cite{liu2024matcher} uses
all-purpose feature matching for one-shot segmentation.
Recent work has also explored training‑free pipelines for segmentation,
including low‑data regimes such as fine‑grained fungi classification
\cite{cavada2026training} 
\textbf{How Open-V differs.} While FSS-SAM3, VRP-SAM, and related work
target the FSS protocol (single novel class per query, no base-class
inference), Open-V targets the stricter GFSS protocol requiring joint
base-and-novel single-pass inference. Unlike TFM$^2$, which appends a
cache-based adapter to a trained OVSS backbone, Open-V introduces no
trained component at any stage. Unlike LPOSS, which operates on a
pre-specified class list through label propagation, Open-V's class
vocabulary is a runtime argument with natural support for arbitrary category
vocabularies at inference time.
 
\section{Method}
\label{sec:method}
 
\subsection{Problem Formulation}
\label{sec:problem}
 
We follow the strict GFSS protocol of~\cite{tian2022generalized,hossain2024visual}.
A query image $I_q \in \mathbb{R}^{H \times W \times 3}$ must be segmented
into $1 + C_b + C_n$ classes (background, $C_b$ base, $C_n$ novel) in a
single forward pass. Base classes have dense training annotations; novel
classes are specified at test time through a per-class support set
$\mathcal{S}_c = \{(I^k_s, M^k_s)\}^K_{k=1}$ for each $c \in \mathcal{C}_n$,
with $K \in \{1, 5\}$. No novel-class sample is seen at training time.
Open-V eliminates the training stage entirely: every component is a frozen
pretrained foundation model, and the GFSS task is solved at inference
without a single parameter update.
 
\subsection{Semantic Arbitration: Conceptual Framing}
\label{sec:framing}
 
We recast GFSS as \emph{open-vocabulary semantic arbitration}. Each class
$c \in \mathcal{C}_b \cup \mathcal{C}_n$ contributes a per-pixel hypothesis
score drawn from a combination of:
\begin{itemize}[leftmargin=*,topsep=1pt,itemsep=1pt]
  \item A \emph{foundation text prior}: SAM3-PCS returns instance masks
  and calibrated presence scores conditioned on the bare class name.
  \item A \emph{few-shot visual anchor}: a $K$-shot CLIP support centroid,
  applied only to novel classes, reranks SAM3 instances by cosine similarity
  to the support appearance.
\end{itemize}
The arbitration step selects, for each pixel, the class with the highest
arbitrated score. Three things follow from this framing. The vocabulary is
a property of the arbitration inputs, not of any parameter; changing it
requires no retraining. The few-shot signal influences exactly one of the
two evidence sources rather than the arbitration rule itself. The dominant
design choices are calibration and combination, not model fitting.
 
\subsection{Tier 1: SAM3-PCS for Open-Vocabulary Segmentation}
\label{sec:tier1}
 
SAM3's PCS interface accepts a bare text concept and returns an instance set
\begin{equation}
\mathcal{P}_c = \{(m_{c,j},\, s_{c,j},\, b_{c,j})\}^{N_c}_{j=1},\quad
s_{c,j} = \sigma(\ell_{c,j}) \cdot \sigma(\rho_{c,j}),
\label{eq:pcs}
\end{equation}
where $m_{c,j} \in \{0,1\}^{H \times W}$ is an instance mask, $b_{c,j}$ its
bounding box, and $s_{c,j} \in [0,1]$ is the calibrated sigmoid product of
SAM3's per-instance quality logit $\ell_{c,j}$ and presence logit $\rho_{c,j}$.
 
We pass the bare class name (\emph{e.g.}, ``aeroplane'') rather than any
CLIP-style template; empirically, SAM3 returns zero masks for
``a photo of a \{cls\}'' on all tested images, because the model was trained on
bare-concept prompts and the templated variant falls outside its calibration.
The SAM3 backbone is computed once per query; iterating over $C_b + C_n$
classes costs one backbone forward pass and $C_b + C_n$ lightweight decoder
passes, as the backbone state is cached across classes.
 
\textbf{Presence floor and class-level mask.} We apply a presence floor
$\tau_\text{PCS} = 0.20$ to filter low-confidence instances. Defining the
post-floor index set $\mathcal{J}_c = \{j : \tilde{s}_{c,j} \geq \tau_\text{PCS}\}$,
the class-level semantic mask and class score are:
\begin{equation}
M_c = \bigcup_{j \in \mathcal{J}_c} m_{c,j},\qquad
\sigma_c = \max_{j \in \mathcal{J}_c} \tilde{s}_{c,j}.
\label{eq:union}
\end{equation}
If $\mathcal{J}_c = \emptyset$, the class is treated as absent in this
query: $M_c = \mathbf{0}$ and $\sigma_c = -\infty$.
 
\subsection{K-Shot CLIP Support-Centroid Rerank}
\label{sec:centroid}
 
SAM3's released image processor accepts text and geometric prompts only; no
exemplar-mask prompt interface is exposed at inference time (see
Sec.~\ref{sec:negative} for the spatial-concatenation workaround we
evaluated and its negative result). We therefore inject the $K$-shot signal
through CLIP's joint vision-language space as a post-hoc rerank.
 
\textbf{Per-class visual centroid.} For each novel class $c \in \mathcal{C}_n$,
define the masked-foreground operator $\phi(\mathbf{I}, \mathbf{M}) =
\mathbf{I} \odot \mathbf{M}$ and the L$_2$-normalised CLIP image embedder
$f_\text{CLIP}(\cdot) = \text{CLIP}_\text{img}(\cdot) /
\|\text{CLIP}_\text{img}(\cdot)\|_2$. The support centroid is:
\begin{equation}
\mathbf{v}_c = \frac{\sum_{k=1}^K f_\text{CLIP}(\phi(I^k_s, M^k_s))}
{\left\|\sum_{k=1}^K f_\text{CLIP}(\phi(I^k_s, M^k_s))\right\|_2}
\in \mathbb{R}^d,\quad \|\mathbf{v}_c\|_2 = 1.
\label{eq:centroid}
\end{equation}
$\mathbf{v}_c$ is computed once per evaluation run and shared across all
queries; CLIP is never re-invoked on the support side.
Concretely, $\phi(I, M)$ crops the image to the tight bounding
box of $M$, zeros pixels outside $M$ within that crop, and
passes the result to the frozen CLIP image encoder, which
resizes it to $224{\times}224$ internally; query instances
$m_{c,j}$ are embedded identically using $\phi(I_q, m_{c,j})$.
\textbf{Per-instance score rerank.} At inference, for each novel-class
SAM3 instance $(m_{c,j}, s_{c,j}) \in \mathcal{P}_c$ we extract the
corresponding query region and embed it:
\begin{equation}
\mathbf{e}_{c,j} = f_\text{CLIP}\!\left(\phi(I_q, m_{c,j})\right) \in \mathbb{R}^d,
\quad \|\mathbf{e}_{c,j}\|_2 = 1,
\label{eq:embed}
\end{equation}
so $\mathbf{e}_{c,j}^\top \mathbf{v}_c = \cos(\mathbf{e}_{c,j}, \mathbf{v}_c)
\in [-1,1]$ measures cosine similarity between the query region and the class
centroid. The reranked score linearly fuses SAM3 confidence with support-conditioned 
CLIP similarity, with $\alpha$ controlling the contribution of each evidence source:
\begin{equation}
\tilde{s}_{c,j} = (1 - \alpha)\, s_{c,j} + \alpha\, \mathbf{e}_{c,j}^\top \mathbf{v}_c,
\qquad \alpha \in [0,1].
\label{eq:rerank}
\end{equation}
Setting $\alpha = 0$ recovers pure SAM3-PCS; $\alpha = 1$ relies on the
K-shot centroid alone. \textbf{$\alpha$ is the only continuous
hyperparameter Open-V introduces}; we fix $\alpha = 0.5$ throughout.
Base-class instances are not reranked: their text prompts are pretraining-strong
and the strict GFSS protocol provides no base support set from which to
build a centroid.
 
\subsection{Tier 2: Boundary-Band Refinement}
\label{sec:tier2}
 
Tier-1 masks are accurate in object interiors but may be imprecise along
boundaries, since PCS optimises for instance discovery. We include an
optional second SAM3 call that ports the boundary-band refinement
of~\cite{liu2022intermediate}: for each class $c$ with $M_c \neq \mathbf{0}$,
let $\partial M_c$ be the morphological boundary of $M_c$,
$B_c = D_\delta(\partial M_c)$ the $\delta$-pixel dilation (boundary band),
and $b_c^+$ the bounding box of $M_c$ enlarged by $\eta$ pixels. Querying
SAM3 with box prompt $b_c^+$ returns $\hat{M}_c$; we update only the
band-restricted pixels:
\begin{equation}
M_c \leftarrow \left(M_c \setminus B_c\right) \cup \left(B_c \cap \hat{M}_c\right).
\label{eq:band}
\end{equation}
The second SAM3 call reuses the backbone state cached by Tier 1 (no
additional image encoding). We use $\delta = 5$ and $\eta = 8$ pixels.
The class score $\sigma_c$ is left unchanged. Tier-2's empirical
contribution on PASCAL is at the noise floor ($\Delta = +0.02$ HM on a
200-episode probe), consistent with PCS already producing pixel-tight masks
at PASCAL object scales; we retain it for open-vocabulary categories where
Tier-1 localisation may be weaker.
 
\subsection{Per-Pixel Semantic Arbitration}
\label{sec:arbitration}
 
The refined class masks $\{M_c\}_{c \in \mathcal{C}_b \cup \mathcal{C}_n}$
may overlap because SAM3-PCS processes each class independently. Letting
$\mathcal{A}_p = \{c : M_c(p) = 1\}$ denote the set of classes claiming
pixel $p$:
\begin{equation}
\text{Label}(p) = \begin{cases}
\arg\max_{c \in \mathcal{A}_p}\, \sigma_c & \mathcal{A}_p \neq \emptyset,\\
0\, \text{(background)} & \mathcal{A}_p = \emptyset.
\end{cases}
\label{eq:argmax}
\end{equation}
The rule is symmetric in base and novel classes, consistent with the strict
GFSS protocol. The rerank of Eq.~\eqref{eq:rerank} is therefore the only
place where novel classes are treated specially, and it carries all of the
few-shot signal.
 
\subsection{Spatial-Alignment Requirement}
\label{sec:alignment}
 
Open-V inherits a non-obvious requirement from the interaction between the
SAM3 image processor and the strict-protocol GFSS data loader. The
strict-protocol loader applies an aspect-preserving letterbox resize (longest
side to $S$, zero-pad to $S \times S$, ground-truth padded with ignore
index 255). The natural drop-in implementation—re-loading the disk image
and calling a square resize—places object centroids at spatial positions
incompatible with the loader's letterbox, causing silent evaluation failure.
We feed the \emph{same letterboxed image} produced by the loader directly
into SAM3, sharing a single spatial frame. This one-line fix has a large
empirical effect (Sec.~\ref{sec:align_diag}).
 
\section{Experiments}
\label{sec:experiments}
 
\subsection{Setup}
\label{sec:setup}
 
\textbf{Datasets and protocols.}
\textit{PASCAL-5$_i$}: 4-fold cross-validation over 20 PASCAL VOC classes
(15 base + 5 novel per fold). We report per-fold and mean results under
both 1-shot and 5-shot at full validation (${\sim}1450$ episodes per fold),
averaged over five independent runs.
\textit{COCO-20$^i$}: 4-fold; 60 base + 20 novel per fold. Full-val
($40{,}117$ episodes per fold) is infeasible at SAM3 ViT-L's ${\sim}23$\,s
per query on the hardware used (estimated ${\sim}$10 days per fold per run),
so we follow the episode-sampling evaluation protocol adopted by prior
work~\cite{hossain2024visual,tian2022generalized}.
\textit{ADE-OW}: A 26-class held-out subset of ADE-20K constructed by
removing every class whose WordNet noun-synset lemma overlaps the union of
PASCAL VOC (20 classes), COCO-Stuff (181 classes), and ImageNet-1K (1000
labels), as a label-level control for long-tail open-vocabulary evaluation.
The full class list with ADE indices and synset mappings is given in
Appendix~\ref{sec:appendix_ade_ow_construction}, and the subset is fully
reproducible from the stated lemma-overlap criterion.
 
\textbf{Metrics.} Mean intersection-over-union over base classes
(mIoU$_b$), novel classes (mIoU$_n$), and their harmonic mean
$\text{HM} = 2 \cdot \text{mIoU}_b \cdot \text{mIoU}_n / (\text{mIoU}_b
+ \text{mIoU}_n)$. HM is the headline GFSS metric.
 
\textbf{Implementation.} Open-V uses SAM3 ViT-L (\texttt{facebook/sam3})
and CLIP ViT-B/16 (\texttt{openai}), both frozen with no task-specific
parameter updates. Inference runs at $1024 \times 1024$ resolution on a
single L4 GPU (24\,GB). The presence floor is fixed at $\tau_\text{PCS} =
0.20$; the centroid weight at $\alpha = 0.5$ for novel classes (0 for base).
The spatial-alignment fix of Sec.~\ref{sec:alignment} is applied throughout.
 
\subsection{Main Results on PASCAL-5$_i$}
\label{sec:pascal}
 
Table~\ref{tab:pascal} compares Open-V against trained GFSS baselines. The
+17.7 HM gap over the strongest trained baseline (Visual
Prompting~\cite{hossain2024visual}) decomposes into three factors: (i)
SAM3's foundation-scale pretraining already covers the PASCAL class
vocabulary; (ii) the CLIP centroid rerank disambiguates novel classes
without parameter updates; and (iii) spatial alignment (Sec.~\ref{sec:align_diag})
closes the dominant silent failure mode.
The +17.7 HM margin reflects the complementary 
strengths of foundation-scale text priors and the 
Open-V arbitration design; rows marked $\dagger$ 
indicate that Open-V operates over an open 
vocabulary while trained baselines fix their 
class list at training time. The ADE-OW
evaluation (Sec.~\ref{sec:ablations}) further validates
Open-V on a label-disjoint vocabulary outside 
any standard benchmark. 

On in-distribution PASCAL categories the foundation prior is already strong:
with no support set, zero-shot SAM3-PCS reaches 76.5 HM on the matched
split-0 probe, and adding the $K$-shot centroid ($\alpha=0.5$) lifts
full-validation HM to 77.9. A modest in-distribution margin is expected when
the class vocabulary lies inside the foundation prior; the support signal's
value emerges where that prior weakens, widening by $6.6\times$ on the
label-disjoint ADE-OW vocabulary (Sec.~\ref{sec:ablations}). Open-V supplies
this signal without retraining, so it is available exactly when the text
prior is insufficient.
 
Per-fold HM is 77.49/78.89/78.84/76.20 (1-shot) and
77.02/78.52/78.87/75.99 (5-shot) for folds 0–3 (five-run mean). Cross-fold
spread is tight (2.69 pp at 1-shot) compared with 5–10 pp typical of trained
baselines that fit a separate decoder per fold; Open-V shares a single frozen
backbone, so between-fold variability comes only from support sampling.
\begin{table}[t]
\centering
\caption{Strict-GFSS results on PASCAL-5$_i$, mean over 4 folds (full validation, 5 runs per fold unless noted). Bold: best 1-shot HM. $\dagger$: frozen SAM3/CLIP (PASCAL vocabulary seen). $\ddagger$: split-0 200-episode probe.}
\label{tab:pascal}
\setlength{\tabcolsep}{3pt}
\scriptsize
\begin{tabular}{lccccc}
\toprule
Method & Trained? & OV? & mIoU$_b$ & mIoU$_n$ & HM \\
\midrule
BCM~\cite{tian2022generalized} & \checkmark & & 75.0 & 47.0 & 57.8 \\
Vis. Prompting~\cite{hossain2024visual} & \checkmark & & 74.9 & 50.3 & 60.2 \\
\midrule
Open-V ($\alpha{=}0$, no align)$\dagger$ & & \checkmark & 45.6 & 28.9 & 35.3 \\
Open-V (zero-shot, aligned)$\dagger\ddagger$ & & \checkmark & 76.0 & 76.9 & 76.5 \\
Open-V ($\alpha{=}0.5$, 1-shot)$\dagger$ & & \checkmark & 78.4 & 77.5 & \textbf{77.9} \\
Open-V ($\alpha{=}0.5$, 5-shot)$\dagger$ & & \checkmark & 78.5 & 76.9 & 77.6 \\
\bottomrule
\end{tabular}
\end{table}

\begin{figure}[t]
  \centering
  \includegraphics[width=0.82\columnwidth]{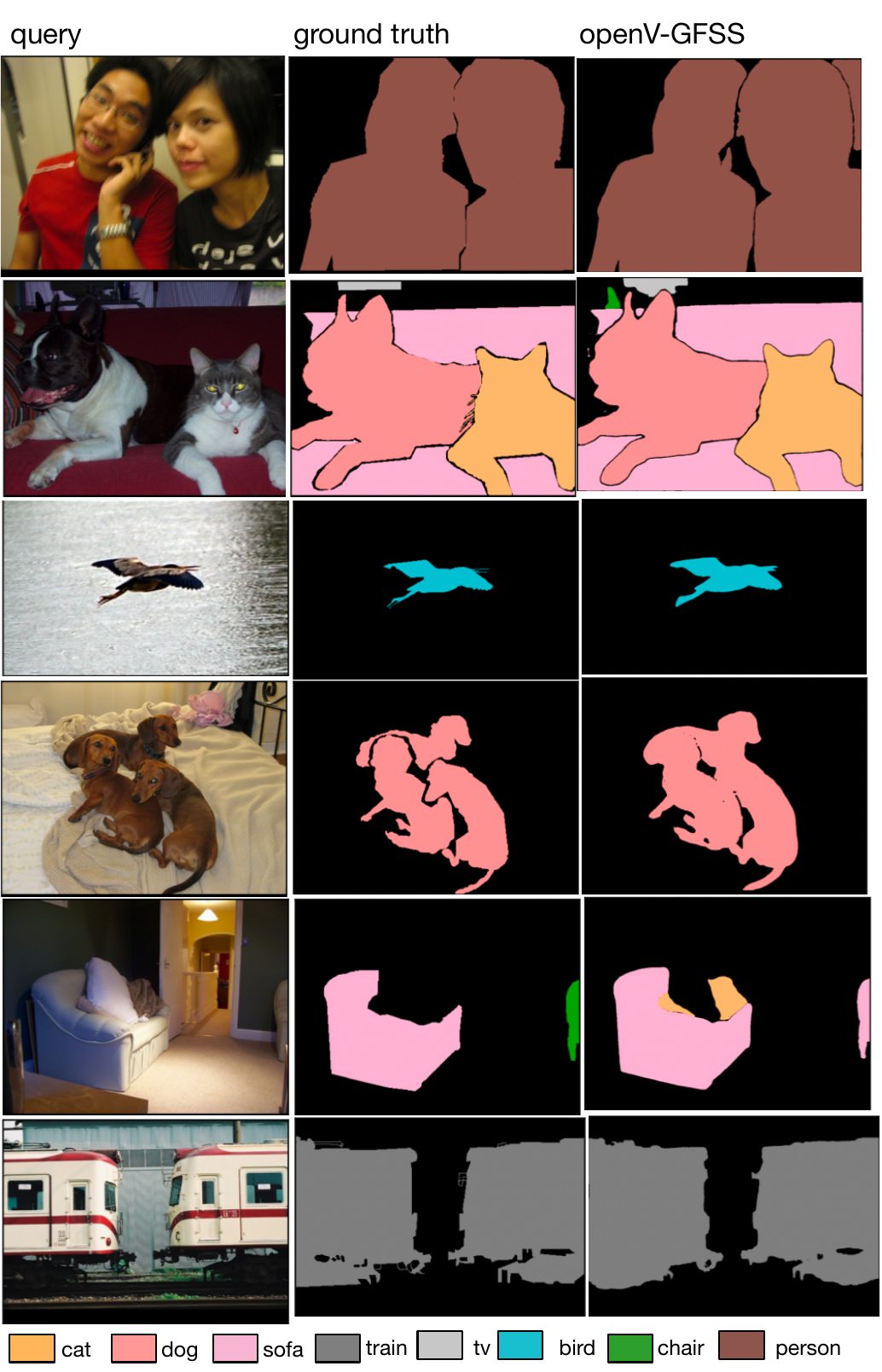}
  \caption{Qualitative strict-GFSS predictions on PASCAL-$5^i$ (input,
  ground truth, \method). Each row mixes base and novel classes in a
  single forward pass: \method recovers thin and articulated novel
  structures (\emph{bird}, \emph{dog}) and large base regions
  (\emph{train}, \emph{person}) without per-split training. Best viewed
  zoomed in.}
  \label{fig:qual_pascal}
\end{figure}
\subsection{Main Results on COCO-20$^i$}
\label{sec:coco}
 
Table~\ref{tab:coco} reports five-run means for all four COCO-20$^i$ folds.
The cross-fold 1-shot mean is HM = 58.17 (mIoU$_b$ = 58.92, mIoU$_n$ =
57.81). The 5-shot cross-fold mean is HM = 58.35—a near-tie with 1-shot
($\leq 0.18$ pp HM difference), consistent with the saturated-vocabulary
reading: additional shots cannot recover SAM3 mass the text prompt already
failed to produce when the class vocabulary is within the foundation
pretraining scope.
 
Cross-fold spread on COCO (${\sim}5.7$ pp HM between splits 1 and 3) is
wider than on PASCAL ($2.7$ pp). This tracks how well each fold's novel
vocabulary maps onto SAM3's text prior. Split-1's novel set consists of
largely common-object vocabulary (bicycle, bus, bear, pizza, etc.), lifting
novel mIoU to 66.49. Splits 2–3 include concept-ambiguous classes (mouse:
animal or input device; hair drier: 0\% detection in both splits), pulling
novel mIoU to 55.16 and 51.95 respectively. This is the foundation
prior-coverage caveat at fold granularity.
 
\begin{table}[t]
\centering
\caption{COCO-20$^i$ results (1-shot and 5-shot), mean over 5 runs per fold.
Protocol matches~\cite{hossain2024visual,tian2022generalized}. $\sigma_\text{HM} \leq 0.20$ pp on splits 0–2, 0.45 pp on split-3.}
\label{tab:coco}
\setlength{\tabcolsep}{5pt}
\small
\begin{tabular}{lccc|ccc}
\toprule
 & \multicolumn{3}{c|}{1-shot} & \multicolumn{3}{c}{5-shot} \\
Fold & mIoU$_b$ & mIoU$_n$ & HM & mIoU$_b$ & mIoU$_n$ & HM \\
\midrule
Split-0 & 59.38 & 57.63 & 58.49 & 59.52 & 57.94 & 58.72 \\
Split-1 & 56.66 & 66.49 & 61.18 & 56.72 & 66.10 & 61.05 \\
Split-2 & 60.19 & 55.16 & 57.57 & 60.21 & 55.27 & 57.63 \\
Split-3 & 59.44 & 51.95 & 55.44 & 60.00 & 52.49 & 55.99 \\
\midrule
\textbf{Mean} & \textbf{58.92} & \textbf{57.81} & \textbf{58.17} & \textbf{59.11} & \textbf{57.95} & \textbf{58.35} \\
\bottomrule
\end{tabular}
\end{table}
 
\subsection{Spatial-Alignment Diagnosis}
\label{sec:align_diag}
 
Table~\ref{tab:alignment} isolates the spatial-alignment confound on PASCAL-5$_i$
split-0, 1-shot, over a 200-episode probe, under two predictor backbones.
For SAM3-PCS, switching from a naïve square resize to the loader's
letterbox lifts HM from 35.79\% to 72.92\%—a $+$37.1 pp shift.
 
To confirm the failure is a property of the foundation-segmenter integration
class rather than of SAM3 specifically, we replicate under the canonical
SAM2-AutoMask + CLIP rerank backbone, matched for Tier-2, $\alpha$, support
sampling, and evaluator. SAM2 + CLIP also degrades catastrophically under
square-stretch (28.61\% to 12.25\%, $\Delta = -16.36$ pp). Two foundation
segmenters of different architecture and different training corpus fail
silently in the same way under the same evaluator mismatch.
 
\textbf{Why this is silent.} The confound is invisible at the level of image
resolution, mask resolution, or foreground pixel fraction. It surfaces
through per-class IoU on spatially contained objects (cat, chair,
pottedplant), whose centroids are displaced by a larger fraction of the
normalised frame than elongated objects (bus, train). The minimal diagnostic
check we recommend: measure the mean object-centroid displacement between
the predictor's spatial transform $T_\text{pred}(I)$ and the evaluator's
$T_\text{eval}(I)$ on a small validation sample; a displacement exceeding a
few percent of the canvas size on small objects signals incompatibility.
 
\begin{table}[t]
\centering
\caption{Spatial-alignment diagnosis. PASCAL-5$_i$ split-0, 1-shot,
200-episode probe. Both backbones drop $>$10 HM pp under square-stretch,
establishing the failure as a class property of foundation-segmenter
integrations.}
\label{tab:alignment}
\setlength{\tabcolsep}{4pt}
\small
\begin{tabular}{llccc}
\toprule
Tier-1 backbone & Preprocess & mIoU$_b$ & mIoU$_n$ & HM \\
\midrule
SAM3-PCS & letterbox & 75.94 & 70.14 & 72.92 \\
SAM3-PCS & cv2.resize & 40.93 & 31.81 & 35.79 \\
SAM2 + CLIP & letterbox & 26.19 & 31.51 & 28.61 \\
SAM2 + CLIP & cv2.resize & 15.35 & 10.19 & 12.25 \\
\midrule
$\Delta$ (SAM3-PCS) & lbox$\to$sq & $-35.01$ & $-38.33$ & $-37.13$ \\
$\Delta$ (SAM2+CLIP) & lbox$\to$sq & $-10.84$ & $-21.32$ & $-16.36$ \\
\bottomrule
\end{tabular}
\end{table}
 
\subsection{Ablation Studies}
\label{sec:ablations}
 
\textbf{Support-centroid weight $\alpha$ (Table~\ref{tab:alpha_pascal}).}
On PASCAL-5$_i$ split-0, sweeping $\alpha$ from 0 to 1 moves HM within a
0.44 pp band (76.11 to 76.55). The contribution of the $K$-shot centroid
over pure text-conditioned inference ($\alpha = 0$) is $\Delta = +0.10$ HM,
of the same order as run-to-run variance. SAM3-PCS calibration is already
strong on PASCAL classes, leaving limited headroom for the visual centroid
to shift the decision boundary.
 
\begin{table}[t]
\centering
\caption{Effect of $\alpha$ (Eq.~\ref{eq:rerank}). PASCAL-5$_i$ split-0,
1-shot, 200-episode probe. Bold: default.}
\label{tab:alpha_pascal}
\small
\begin{tabular}{cccc}
\toprule
$\alpha$ & mIoU$_b$ & mIoU$_n$ & HM \\
\midrule
0.0 & 75.97 & 76.94 & 76.45 \\
0.3 & 75.97 & 76.95 & 76.46 \\
\textbf{0.5} & \textbf{75.94} & \textbf{77.17} & \textbf{76.55} \\
0.7 & 75.88 & 76.33 & 76.11 \\
1.0 & 75.91 & 77.04 & 76.47 \\
\bottomrule
\end{tabular}
\end{table}
 
\textbf{Few-shot signal on a label-disjoint vocabulary (Table~\ref{tab:alpha_ade}).}
Repeating the $\alpha$-sweep on ADE-OW reveals a qualitatively different
picture. The optimal $\alpha$ shifts from 0.5 on PASCAL to 0.7 on ADE-OW;
the sweep span widens from 0.44 pp to 2.90 pp—a factor of 6.6$\times$.
This establishes that the K-shot centroid's contribution scales with the
degree to which the foundation text prior weakens. The collapse at $\alpha=1.0$
($-2.90$ pp from peak) shows that pure-centroid replacement loses the text
prior on classes where SAM3-PCS still segments accurately; the convex
combination over either extreme is empirically motivated by this asymmetry.
 
The per-class breakdown in Table~\ref{tab:alpha_perclass} further resolves
the aggregate into three interaction regimes: \emph{prior-dominant} classes
(flat IoU across $\alpha$, e.g., tower, flag, stairway); \emph{ambiguity-resolving}
classes (IoU peaks at interior $\alpha$, e.g., sconce $41.2 \to 51.0$,
streetlight $65.7 \to 67.6$); and \emph{prior-conflicting} classes (IoU
collapses at $\alpha=1.0$, e.g., chandelier $73.5 \to 44.6$, fan $83.8 \to 73.3$).
The growth of the ambiguity-resolving regime as the foundation prior weakens
is precisely what produces the 6.6$\times$ widening of the sweep span.
 
\begin{table}[t]
\centering
\caption{Effect of $\alpha$ on ADE-OW (26 label-disjoint classes from
ADE-20K, $K=5$, $N=860$ queries). The sweep span widens 6.6$\times$
vs.\ PASCAL, tracking the weakening of the foundation text prior. Bold: ADE-OW peak.}
\label{tab:alpha_ade}
\small
\begin{tabular}{ccc}
\toprule
$\alpha$ & mIoU (\%) & $\Delta$ vs.\ $\alpha{=}0$ \\
\midrule
0.0 & 60.65 & 0.00 \\
0.3 & 60.64 & $-$0.01 \\
0.5 & 60.95 & $+$0.30 \\
\textbf{0.7} & \textbf{61.62} & $+$0.97 \\
1.0 & 58.72 & $-$1.93 \\
\bottomrule
\end{tabular}
\end{table}
 
\begin{table}[t]
\centering
\caption{Per-class IoU (\%) on ADE-OW (representative classes from
each regime; all 26 classes with ADE indices and synonyms in
Appendix~\ref{sec:appendix_ade_ow_construction}).
$n$: query count after $K{=}5$ support removal.}
\label{tab:alpha_perclass}
\setlength{\tabcolsep}{2.5pt}
\scriptsize
\begin{tabular}{lrrrrrr}
\toprule
Class & $\alpha{=}0$ & $\alpha{=}0.3$ & $\alpha{=}0.5$
      & $\alpha{=}0.7$ & $\alpha{=}1.0$ & $n$ \\
\midrule
\multicolumn{7}{l}{\textit{Prior-dominant}} \\
tower        & 93.0 & 93.0 & 93.0 & 93.0 & 92.8 &   9 \\
flag         & 84.7 & 84.8 & 84.8 & 84.7 & 84.7 &  49 \\
stairway     & 80.7 & 80.7 & 80.7 & 80.7 & 80.7 &  44 \\
\midrule
\multicolumn{7}{l}{\textit{Ambiguity-resolving}} \\
streetlight  & 65.7 & 66.2 & 67.2 & 67.6 & 67.9 & 228 \\
basket       & 54.3 & 54.3 & 59.3 & 59.1 & 55.1 &  63 \\
sconce       & 41.2 & 45.8 & 47.3 & 51.0 & 41.7 &  96 \\
\midrule
\multicolumn{7}{l}{\textit{Prior-conflicting}} \\
fan          & 83.8 & 82.2 & 82.6 & 72.6 & 73.3 &  35 \\
chandelier   & 73.5 & 71.6 & 70.3 & 68.7 & 44.6 &  51 \\
lamp         & 72.5 & 72.5 & 72.8 & 71.1 & 61.5 & 274 \\
\bottomrule
\end{tabular}
\end{table}
 
\subsection{Runtime and Scalability}
\label{sec:runtime}
 
Table~\ref{tab:runtime} reports per-query wall time on a single L4 GPU.
SAM3's backbone is invoked once per query and cached; the PCS decoder is
invoked once per class on that cached state, so cost grows linearly in
class count. The CLIP rerank over $N_c \sim 1$–8 instances per novel class
is negligible against the PCS-decoder cost. The support centroid is computed
once per evaluation run and adds no per-query cost.
 
At 500 classes, the same architectural shape predicts ${\sim}$2 minutes per
query, dominated by 500 PCS-decoder passes. Two practical mitigations are
available: a cheap CLIP zero-shot whole-image pre-filter to prune classes,
and batching the independent per-class decoder calls. We pursue neither in
this paper but note both as routes to bring large-vocabulary Open-V into
interactive time budgets.
 
\begin{table}[t]
\centering
\caption{Per-query wall time, L4 GPU ($1024^2$). One backbone pass is
amortised; per-class decoder cost grows linearly.}
\label{tab:runtime}
\small
\begin{tabular}{lccc}
\toprule
Benchmark & $C_b+C_n$ & s/query & 500-class proj. \\
\midrule
PASCAL-5$_i$ & 20 & $\sim$6 & --- \\
COCO-20$^i$ & 80 & $\sim$23 & $\sim$120 s \\
\bottomrule
\end{tabular}
\end{table}
 
\subsection{Negative Result: Spatial-Concatenation Exemplar Prompting}
\label{sec:negative}
 
We evaluated a dual-stream variant in which novel classes are segmented via
SAM3-PCS on a [query | support] canvas, following the spatial-concatenation
strategy of FSS-SAM3~\cite{tsai2026fsssam3}. The intent was to provide a
visual exemplar anchor within the SAM3 input space, by analogy to
in-context image prompting. On split-0 1-shot, this variant did not
improve over the CLIP-centroid rerank once spatial alignment was corrected.
 
Inspecting SAM3's processor API clarifies the result: the released
\texttt{Sam3Processor} accepts text and geometric (point/box) prompts
only; no exemplar-mask prompt slot is populated at the image-level in the
released code. A canvas + text query thus reduces to running PCS on a
noisier, larger image, with support-side instances filtered out in
post-processing. We report this as a negative result. Integrating SAM3
with image-exemplar prompting will require a model-side interface, not a
canvas-side workaround. This finding echoes the observation in
FSS-SAM3~\cite{tsai2026fsssam3} that negative prompts can cause prediction
collapse through conflicting spatial signals—both results point to
limitations in how current foundation models handle competing semantic
inputs.
 
\section{Limitations and Open Questions}
\label{sec:limits}
 
\textbf{Foundation-pretraining coverage.} PASCAL-5$_i$ and COCO-20$^i$
classes are plausibly covered by SAM3's text-encoder pretraining and
SA-1B-style mask pretraining. Following the annotation convention of
VRP-SAM~\cite{sun2024vrpsam}, we mark foundation-based rows in
Table~\ref{tab:pascal} with $\dagger$. We therefore scope the $+$17.7 HM
result as evidence that training-free arbitration of large pretrained priors
can match or surpass trained GFSS baselines on in-distribution categories,
with the support centroid supplying the novel-class disambiguation that a
text prompt alone does not. The role of that signal grows as the prior
weakens: our ADE-OW ablation (Sec.~\ref{sec:ablations}) shows a 6.6$\times$
widening on a label-disjoint vocabulary, and a larger benchmark disjoint from
PASCAL VOC, COCO-Stuff, and ImageNet-1K is the natural next step for the
community.
 
\textbf{No exemplar-mask API in released SAM3.} The released SAM3 image
processor exposes text and box prompts only; the spatial-concatenation
workaround underperforms the CLIP-centroid rerank. A model-side exemplar-mask
interface would likely strengthen the K-shot signal on fine-grained classes
whose text concepts under-cover the visual category.
 
\textbf{Union and class-level arbitration in crowded scenes.} The
class-level scalar $\sigma_c = \max_j \tilde{s}_{c,j}$ is set by the
strongest instance only. In dense multi-class scenes, a single strong
instance can lend its score to weaker same-class instances at distant pixels
via the union. Resolving this requires per-pixel confidence maps, which the
current SAM3 API does not expose.
 
\section{Conclusion}
\label{sec:conclusion}
 
We presented Open-V, a training-free open-vocabulary pipeline for Generalized
Few-Shot Semantic Segmentation, reframed as open-vocabulary semantic
arbitration over per-class hypothesis sources. By coupling SAM3-PCS with a
$K$-shot CLIP support-centroid rerank and adjudicating the two priors through
a calibrated per-pixel arg-max, Open-V matches and surpasses trained GFSS
baselines on PASCAL-5$_i$ and COCO-20$^i$ under the strict protocol, with no
GFSS-specific training, no fine-tuning, and no support-image augmentation.
 
Three findings accompany the segmentation results as methodological
contributions. First, the few-shot signal's contribution to GFSS is
calibration-dependent: on in-distribution PASCAL categories, where the text
prior already covers the vocabulary, the K-shot centroid yields a small
margin, whereas on a label-disjoint vocabulary the same centroid's
contribution widens by 6.6$\times$, tracking the weakening of the foundation
text prior.
Second, spatial alignment between predictor and evaluator is a dominant
silent failure mode that can shift harmonic-mean mIoU by more than 37
percentage points and replicates across predictor backbones—we characterise
it and provide a minimal diagnostic check. Third, spatial-concatenation
exemplar prompting, a natural route to injecting few-shot evidence, fails in
the absence of a model-side exemplar-mask interface, and we document this
as a negative result that delineates the current limits of canvas-side
workarounds.
 
Open-V suggests a broader principle: as foundation models grow stronger,
the scarce resource in few-shot recognition shifts from learned
representations to calibrated arbitration among frozen priors. Designing
arbitration rules that are robust to prior-coverage heterogeneity—and
evaluation protocols that are sensitive to this heterogeneity—is the key
challenge going forward.
 
{
    \small
    \bibliographystyle{ieeenat_fullname}
    \bibliography{main}

\begin{thebibliography}{39}
\providecommand{\natexlab}[1]{#1}
\providecommand{\url}[1]{\texttt{#1}}
\expandafter\ifx\csname urlstyle\endcsname\relax
  \providecommand{\doi}[1]{doi: #1}\else
  \providecommand{\doi}{doi: \begingroup \urlstyle{rm}\Url}\fi

\bibitem[Barsellotti et~al.(2024{\natexlab{a}})Barsellotti, Amoroso, Baraldi,
  and Cucchiara]{barsellotti2024fossil}
Luca Barsellotti, Roberto Amoroso, Lorenzo Baraldi, and Rita Cucchiara.
\newblock {Fossil: Free Open-Vocabulary Semantic Segmentation through Synthetic
  References Retrieval}.
\newblock In \emph{IEEE/CVF Winter Conference on Applications of Computer
  Vision (WACV)}, pages 1464--1473, 2024{\natexlab{a}}.

\bibitem[Barsellotti et~al.(2024{\natexlab{b}})Barsellotti, Amoroso, Cornia,
  Baraldi, and Cucchiara]{barsellotti2024training}
Luca Barsellotti, Roberto Amoroso, Marcella Cornia, Lorenzo Baraldi, and Rita
  Cucchiara.
\newblock {Training-Free Open-Vocabulary Segmentation with Offline
  Diffusion-Augmented Prototype Generation}.
\newblock In \emph{IEEE/CVF Conference on Computer Vision and Pattern
  Recognition (CVPR)}, pages 3689--3698, 2024{\natexlab{b}}.

\bibitem[Carion et~al.(2026)Carion, Gustafson, Hu, Debnath, Hu, Suris, Ryali,
  Alwala, Khedr, Huang, et~al.]{carion2026sam3}
Nicolas Carion, Laura Gustafson, Yuan-Ting Hu, Shoubhik Debnath, Ronghang Hu,
  Didac Suris, Chaitanya Ryali, Kalyan~Vasudev Alwala, Haitham Khedr, Andrew
  Huang, et~al.
\newblock {SAM 3: Segment Anything with Concepts}.
\newblock In \emph{International Conference on Learning Representations
  (ICLR)}, 2026.

\bibitem[Cavada et~al.(2026)Cavada, Pelosin, and Faggi]{cavada2026training}
Sebastian Cavada, Francesco Pelosin, and Lapo Faggi.
\newblock Training-free fine-grained semantic segmentations in low data
  regimes: A fungitastic baseline.
\newblock 2026.
\newblock Accepted at the 13th Workshop on Fine-Grained Visual Categorization,
  CVPR 2026.

\bibitem[Cha et~al.(2023)Cha, Mun, and Roh]{cha2023learning}
Junbum Cha, Jonghwan Mun, and Byungseok Roh.
\newblock {Learning to Generate Text-Grounded Mask for Open-World Semantic
  Segmentation from Only Image-Text Pairs}.
\newblock In \emph{IEEE/CVF Conference on Computer Vision and Pattern
  Recognition (CVPR)}, pages 11165--11174, 2023.

\bibitem[Chen et~al.(2018)Chen, Papandreou, Kokkinos, Murphy, and
  Yuille]{chen2018deeplab}
Liang-Chieh Chen, George Papandreou, Iasonas Kokkinos, Kevin Murphy, and
  Alan~L. Yuille.
\newblock {DeepLab: Semantic Image Segmentation with Deep Convolutional Nets,
  Atrous Convolution, and Fully Connected CRFs}.
\newblock \emph{IEEE Transactions on Pattern Analysis and Machine Intelligence
  (TPAMI)}, 2018.

\bibitem[Cho et~al.(2024)Cho, Shin, Hong, Arnab, Seo, and Kim]{cho2024cat}
Seokju Cho, Heeseong Shin, Sunghwan Hong, Anurag Arnab, Paul~Hongsuck Seo, and
  Seungryong Kim.
\newblock {CAT-Seg: Cost Aggregation for Open-Vocabulary Semantic
  Segmentation}.
\newblock In \emph{IEEE/CVF Conference on Computer Vision and Pattern
  Recognition (CVPR)}, pages 4113--4123, 2024.

\bibitem[Ding et~al.(2023)Ding, Wang, and Tu]{ding2023open}
Zheng Ding, Jieke Wang, and Zhuowen Tu.
\newblock {Open-Vocabulary Universal Image Segmentation with MaskCLIP}.
\newblock In \emph{International Conference on Learning Representations
  (ICLR)}, 2023.

\bibitem[Fan et~al.(2022)Fan, Pei, Tai, and Tang]{fan2022self}
Qi Fan, Wenjie Pei, Yu-Wing Tai, and Chi-Keung Tang.
\newblock {Self-Support Few-Shot Semantic Segmentation}.
\newblock In \emph{European Conference on Computer Vision (ECCV)}, 2022.

\bibitem[Geng et~al.(2024)]{geng2024enhancing}
Xinwei Geng et~al.
\newblock {Enhancing Generalized Few-Shot Semantic Segmentation via Effective
  Knowledge Transfer}.
\newblock In \emph{AAAI Conference on Artificial Intelligence (AAAI)}, 2024.

\bibitem[Hossain et~al.(2024)]{hossain2024visual}
Mir Rayat~Imtiaz Hossain et~al.
\newblock {Visual Prompting for Generalized Few-Shot Segmentation: A
  Multi-Scale Approach}.
\newblock In \emph{IEEE/CVF Conference on Computer Vision and Pattern
  Recognition (CVPR)}, 2024.

\bibitem[Kirillov et~al.(2023)Kirillov, Mintun, Ravi, Mao, Rolland, Gustafson,
  Xiao, Whitehead, Berg, Lo, Dollár, and Girshick]{kirillov2023segment}
Alexander Kirillov, Eric Mintun, Nikhila Ravi, Hanzi Mao, Chloe Rolland, Laura
  Gustafson, Tete Xiao, Spencer Whitehead, Alexander~C. Berg, Wan-Yen Lo, Piotr
  Dollár, and Ross Girshick.
\newblock {Segment Anything}.
\newblock In \emph{IEEE/CVF International Conference on Computer Vision
  (ICCV)}, 2023.

\bibitem[Lang et~al.(2022)Lang, Cheng, Tu, and Han]{lang2022learning}
Chunbo Lang, Gong Cheng, Binfei Tu, and Junwei Han.
\newblock {Learning What Not to Segment: A New Perspective on Few-Shot
  Segmentation}.
\newblock In \emph{IEEE/CVF Conference on Computer Vision and Pattern
  Recognition (CVPR)}, 2022.

\bibitem[Liang et~al.(2023)Liang, Wu, Dai, Li, Zhao, Zhang, Zhang, Vajda, and
  Marculescu]{liang2023ovseg}
Feng Liang, Bichen Wu, Xiaoliang Dai, Kunpeng Li, Yinan Zhao, Hang Zhang,
  Peizhao Zhang, Peter Vajda, and Diana Marculescu.
\newblock {Open-Vocabulary Semantic Segmentation with Mask-Adapted CLIP}.
\newblock In \emph{IEEE/CVF Conference on Computer Vision and Pattern
  Recognition (CVPR)}, 2023.

\bibitem[Liu et~al.(2023)]{liu2023cafs}
Lan Liu et~al.
\newblock {Class-Agnostic Few-Shot Object Counting and Segmentation via
  Cross-Attention Networks}.
\newblock In \emph{AAAI Conference on Artificial Intelligence (AAAI)}, 2023.

\bibitem[Liu et~al.(2024)Liu, Zhu, Li, Chen, Wang, and Shen]{liu2024matcher}
Yang Liu, Muzhi Zhu, Hengtao Li, Hao Chen, Xinlong Wang, and Chunhua Shen.
\newblock {Matcher: Segment Anything with One Shot Using All-Purpose Feature
  Matching}.
\newblock In \emph{International Conference on Learning Representations
  (ICLR)}, 2024.

\bibitem[Liu et~al.(2022)]{liu2022intermediate}
Yuanwei Liu et~al.
\newblock {Intermediate Prototype Mining Transformer for Few-Shot Semantic
  Segmentation}.
\newblock In \emph{Advances in Neural Information Processing Systems
  (NeurIPS)}, 2022.

\bibitem[Luo et~al.(2023)Luo, Bao, Wu, He, and Li]{luo2023segclip}
Huaishao Luo, Junwei Bao, Youzheng Wu, Xiaodong He, and Tianrui Li.
\newblock {SegCLIP: Patch Aggregation with Learnable Centers for
  Open-Vocabulary Semantic Segmentation}.
\newblock In \emph{International Conference on Machine Learning (ICML)}, pages
  23033--23044, 2023.

\bibitem[Min et~al.(2021)Min, Kang, and Cho]{min2021hypercorrelation}
Juhong Min, Dahyun Kang, and Minsu Cho.
\newblock {Hypercorrelation Squeeze for Few-Shot Segmentation}.
\newblock In \emph{IEEE/CVF International Conference on Computer Vision
  (ICCV)}, 2021.

\bibitem[Oquab et~al.(2024)Oquab, Darcet, Moutakanni, Vo, Szafraniec, Khalidov,
  Fernandez, Haziza, Massa, El-Nouby, et~al.]{oquab2024dinov2}
Maxime Oquab, Timothée Darcet, Théo Moutakanni, Huy~V. Vo, Marc Szafraniec,
  Vasil Khalidov, Pierre Fernandez, Daniel Haziza, Francisco Massa, Alaaeldin
  El-Nouby, et~al.
\newblock {DINOv2: Learning Robust Visual Features Without Supervision}.
\newblock \emph{Transactions on Machine Learning Research (TMLR)}, 2024.

\bibitem[Peng et~al.(2023)Peng, Tian, Wu, Wang, Liu, Su, and
  Jia]{peng2023hierarchical}
Bohao Peng, Zhuotao Tian, Xiaoyang Wu, Chengyao Wang, Shu Liu, Jingyong Su, and
  Jiaya Jia.
\newblock {Hierarchical Dense Correlation Distillation for Few-Shot
  Segmentation}.
\newblock In \emph{IEEE/CVF Conference on Computer Vision and Pattern
  Recognition (CVPR)}, 2023.

\bibitem[Radford et~al.(2021)Radford, Kim, Hallacy, Ramesh, Goh, Agarwal,
  Sastry, Askell, Mishkin, Clark, Krueger, and Sutskever]{radford2021learning}
Alec Radford, Jong~Wook Kim, Chris Hallacy, Aditya Ramesh, Gabriel Goh,
  Sandhini Agarwal, Girish Sastry, Amanda Askell, Pamela Mishkin, Jack Clark,
  Gretchen Krueger, and Ilya Sutskever.
\newblock {Learning Transferable Visual Models from Natural Language
  Supervision}.
\newblock In \emph{International Conference on Machine Learning (ICML)}, 2021.

\bibitem[Ravi et~al.(2025)Ravi, Gabeur, Hu, Hu, Ryali, Ma, Khedr, Rädle,
  Rolland, Gustafson, et~al.]{ravi2024sam2}
Nikhila Ravi, Valentin Gabeur, Yuan-Ting Hu, Ronghang Hu, Chaitanya Ryali,
  Tengyu Ma, Haitham Khedr, Roman Rädle, Chloe Rolland, Laura Gustafson,
  et~al.
\newblock {SAM 2: Segment Anything in Images and Videos}.
\newblock In \emph{International Conference on Learning Representations
  (ICLR)}, 2025.

\bibitem[Stojnić et~al.(2025)Stojnić, Kalantidis, Matas, and
  Tolias]{stojnic2025lposs}
Vladan Stojnić, Yannis Kalantidis, Jiri Matas, and Giorgos Tolias.
\newblock {LPOSS: Label Propagation Over Patches and Pixels for Open-Vocabulary
  Semantic Segmentation}.
\newblock \emph{arXiv preprint arXiv:2503.19777}, 2025.

\bibitem[Sun et~al.(2024)Sun, Chen, Zhang, Zhang, Chen, Zhang, Ding, Wang, and
  Li]{sun2024vrpsam}
Yanpeng Sun, Jiahui Chen, Shan Zhang, Xinyu Zhang, Qiang Chen, Gang Zhang,
  Errui Ding, Jingdong Wang, and Zechao Li.
\newblock {VRP-SAM: SAM with Visual Reference Prompt}.
\newblock In \emph{IEEE/CVF Conference on Computer Vision and Pattern
  Recognition (CVPR)}, 2024.

\bibitem[Tian et~al.(2020)Tian, Zhao, Shu, Yang, Li, and Jia]{tian2020prior}
Zhuotao Tian, Hengshuang Zhao, Michelle Shu, Zhicheng Yang, Ruiyu Li, and Jiaya
  Jia.
\newblock {Prior Guided Feature Enrichment Network for Few-Shot Segmentation}.
\newblock \emph{IEEE Transactions on Pattern Analysis and Machine Intelligence
  (TPAMI)}, 44\penalty0 (2):\penalty0 1050--1065, 2020.

\bibitem[Tian et~al.(2022)Tian, Zhao, Shu, Yang, Li, and
  Jia]{tian2022generalized}
Zhuotao Tian, Hengshuang Zhao, Michelle Shu, Zhicheng Yang, Ruiyu Li, and Jiaya
  Jia.
\newblock {Generalized Few-Shot Semantic Segmentation}.
\newblock In \emph{IEEE/CVF Conference on Computer Vision and Pattern
  Recognition (CVPR)}, 2022.

\bibitem[Tsai et~al.(2026)Tsai, Lin, and Wang]{tsai2026fsssam3}
Yi-Jen Tsai, Yen-Yu Lin, and Chien-Yao Wang.
\newblock {Few-Shot Semantic Segmentation Meets SAM3}.
\newblock \emph{arXiv preprint arXiv:2604.05433}, 2026.

\bibitem[Wang et~al.(2019)Wang, Liew, Zou, Zhou, and Feng]{wang2019panet}
Kaixin Wang, Jun~Hao Liew, Yingtian Zou, Daquan Zhou, and Jiashi Feng.
\newblock {PANet: Few-Shot Image Semantic Segmentation with Prototype
  Alignment}.
\newblock In \emph{IEEE/CVF International Conference on Computer Vision
  (ICCV)}, 2019.

\bibitem[Wang et~al.(2023)Wang, Luo, and Zhang]{wang2023focus}
Yuan Wang, Naisong Luo, and Tianzhu Zhang.
\newblock Focus on query: Adversarial mining transformer for few-shot
  segmentation.
\newblock \emph{Advances in neural information processing systems},
  36:\penalty0 31524--31542, 2023.

\bibitem[Xie et~al.(2026)Xie, He, Fan, Zhang, Cheng, and Liu]{xie2026make}
Guohuan Xie, Xin He, Dingying Fan, Le Zhang, Ming-Ming Cheng, and Yun Liu.
\newblock Make it up: Fake images, real gains in generalized few-shot semantic
  segmentation.
\newblock \emph{arXiv preprint arXiv:2603.27206}, 2026.

\bibitem[Xu et~al.(2022)Xu, Zhang, Wei, Lin, Cao, Hu, and Bai]{xu2022simseg}
Mengde Xu, Zheng Zhang, Fangyun Wei, Yutong Lin, Yue Cao, Han Hu, and Xiang
  Bai.
\newblock {A Simple Baseline for Open-Vocabulary Semantic Segmentation with
  Pre-Trained Vision-Language Model}.
\newblock In \emph{European Conference on Computer Vision (ECCV)}, pages
  736--753, 2022.

\bibitem[Xu et~al.(2023{\natexlab{a}})Xu, Zhang, Wei, Hu, and Bai]{xu2023san}
Mengde Xu, Zheng Zhang, Fangyun Wei, Han Hu, and Xiang Bai.
\newblock {Side Adapter Network for Open-Vocabulary Semantic Segmentation}.
\newblock In \emph{IEEE/CVF Conference on Computer Vision and Pattern
  Recognition (CVPR)}, pages 2945--2954, 2023{\natexlab{a}}.

\bibitem[Xu et~al.(2023{\natexlab{b}})Xu, Zhao, Lin, and Long]{xu2023self}
Qianxiong Xu, Wenting Zhao, Guosheng Lin, and Cheng Long.
\newblock {Self-Calibrated Cross Attention Network for Few-Shot Segmentation}.
\newblock In \emph{IEEE/CVF International Conference on Computer Vision
  (ICCV)}, 2023{\natexlab{b}}.

\bibitem[Xu et~al.(2024)Xu, Lin, Loy, Long, Li, and Zhao]{xu2024eliminating}
Qianxiong Xu, Guosheng Lin, Chen~Change Loy, Cheng Long, Ziyue Li, and Rui
  Zhao.
\newblock {Eliminating Feature Ambiguity for Few-Shot Segmentation}.
\newblock In \emph{European Conference on Computer Vision (ECCV)}, 2024.

\bibitem[Xu et~al.(2025)Xu, Zhu, Liu, Lin, Long, Li, and Zhao]{xu2025fssam}
Qianxiong Xu, Lanyun Zhu, Xuanyi Liu, Guosheng Lin, Cheng Long, Ziyue Li, and
  Rui Zhao.
\newblock {Unlocking the Power of SAM 2 for Few-Shot Segmentation}.
\newblock In \emph{International Conference on Machine Learning (ICML)}, 2025.

\bibitem[Zhang et~al.(2022)Zhang, Sun, Yang, and Chen]{zhang2022feature}
Jian-Wei Zhang, Yifan Sun, Yi Yang, and Wei Chen.
\newblock {Feature-Proxy Transformer for Few-Shot Segmentation}.
\newblock In \emph{Advances in Neural Information Processing Systems
  (NeurIPS)}, 2022.

\bibitem[Zhao et~al.(2017)Zhao, Puig, Zhou, Fidler, and Torralba]{zhao2017open}
Hang Zhao, Xavier Puig, Bolei Zhou, Sanja Fidler, and Antonio Torralba.
\newblock {Open Vocabulary Scene Parsing}.
\newblock In \emph{IEEE/CVF International Conference on Computer Vision
  (ICCV)}, 2017.

\bibitem[Zhuo et~al.(2025)Zhuo, Bessinger, Wang, Khosravan, Li, and
  Kang]{zhuo2025tfm2}
Yaoxin Zhuo, Zachary Bessinger, Lichen Wang, Naji Khosravan, Baoxin Li, and
  Sing~Bing Kang.
\newblock {TFM$^2$: Training-Free Mask Matching for Open-Vocabulary Semantic
  Segmentation}.
\newblock In \emph{IEEE/CVF Winter Conference on Applications of Computer
  Vision (WACV)}, 2025.

\end{thebibliography}
}

\clearpage
\appendix
\setcounter{table}{0}
\setcounter{figure}{0}
\renewcommand{\thetable}{S\arabic{table}}
\renewcommand{\thefigure}{S\arabic{figure}}

\section{Alignment-Diagnostic Reference Implementation}
\label{sec:appendix_alignment}

This appendix supports the methodological recommendation at the end
of \S\ref{sec:alignment} (``predictor--loader transform
equivalence''). We give an unambiguous specification of the two
transforms compared in Table~\ref{tab:alignment}, a centroid-displacement
diagnostic that surfaces the failure mode on a small
sample of any val set, and a worked numerical example.

\subsection{Transform specification}
\label{sec:appendix_alignment_spec}

Both transforms map a disk image
$\mathbf{I}\!\in\!\mathbb{R}^{H\times W\times 3}$ to a fixed-size
canvas $\mathbf{I}'\!\in\!\mathbb{R}^{S\times S\times 3}$ with
$S\!=\!1024$ in our experiments. They differ only in how aspect ratio
is handled.

\paragraph{Letterbox (ours, matching the GFSS loader).}
Compute $r\!=\!S/\max(H,W)$, resize to $(\lceil rH\rceil,\lceil rW\rceil)$,
zero-pad symmetrically to $S\!\times\!S$. The pad offset is
$p_y\!=\!\lfloor (S-rH)/2\rfloor$ (top) and
$p_x\!=\!\lfloor (S-rW)/2\rfloor$ (left). A pixel at original
coordinates $(y,x)$ maps to
\begin{equation}
  T_{\mathrm{letterbox}}(y,x)
  \;=\;\bigl(p_y + r\,y,\; p_x + r\,x\bigr).
  \label{eq:letterbox}
\end{equation}

\paragraph{Square stretch (failure mode).}
Apply \texttt{cv2.resize(I, (S, S))} directly. The two axes use
independent scales $r_y\!=\!S/H$, $r_x\!=\!S/W$:
\begin{equation}
  T_{\mathrm{square}}(y,x) \;=\; (r_y\,y,\; r_x\,x).
  \label{eq:square}
\end{equation}

These agree only when $H\!=\!W$; for any other aspect ratio, the
square-stretch transform displaces every off-centre pixel. The
ground-truth mask is letterboxed by the loader regardless, so a
prediction made under $T_{\mathrm{square}}$ is scored against a mask
in a different coordinate frame.

\subsection{Centroid-displacement diagnostic}
\label{sec:appendix_alignment_diag}

The minimum check we recommend before trusting any
foundation-segmenter integration is the per-object centroid
displacement between the two transforms, expressed as a fraction of
the canvas. Pseudocode:

\begin{lstlisting}
def centroid_displacement(I_shape, S, mask):
    """
    I_shape: (H, W) of the original disk image
    S:       canvas size used by both pipelines (e.g. 1024)
    mask:    HxW binary object mask in original coords
    Returns: ||T_letterbox(c) - T_square(c)||_2 / S
             where c is the mask centroid.
    """
    H, W = I_shape
    ys, xs = np.where(mask)
    if len(ys) == 0:
        return 0.0
    cy, cx = ys.mean(), xs.mean()

    r = S / max(H, W)
    py = (S - r * H) / 2
    px = (S - r * W) / 2

    ly, lx = py + r * cy, px + r * cx
    sy, sx = (S / H) * cy, (S / W) * cx

    return np.hypot(ly - sy, lx - sx) / S
\end{lstlisting}

The diagnostic returns a scalar in $[0,1]$: zero means the two
transforms place the centroid identically; any value above a few
percent indicates a coordinate-frame mismatch that will silently
suppress IoU on the affected objects. As a rule of thumb, a mean
displacement above $0.02$ on a 50-image sample, computed
class-by-class, is sufficient evidence that the predictor and the
loader are not aligned; the per-class breakdown then reveals which
object scales are most affected.

\subsection{Worked numerical example}
\label{sec:appendix_alignment_example}

We illustrate the diagnostic on a single image with the modal
PASCAL VOC aspect ratio. Take $H\!=\!375$, $W\!=\!500$,
$S\!=\!1024$. Then $r\!=\!1024/500\!=\!2.048$,
$p_y\!=\!(1024-768)/2\!=\!128$, $p_x\!=\!0$, $r_y\!=\!2.731$,
$r_x\!=\!2.048$.

\begin{table}[t]
  \centering
  \footnotesize
  \resizebox{\columnwidth}{!}{%
  \begin{tabular}{@{}lcccc@{}}
    \toprule
    Centroid $(y,x)$ in original
      & $T_{\mathrm{letterbox}}$
      & $T_{\mathrm{square}}$
      & $\Delta$ (px)
      & $\Delta/S$\\
    \midrule
    image centre $(187, 250)$       & $(511, 512)$    & $(511, 512)$    & $0.3$  & $<\!0.001$\\
    upper-left object $(60, 80)$    & $(251, 164)$    & $(164, 164)$    & $87.3$ & $0.085$\\
    upper-right object $(60, 420)$  & $(251, 860)$    & $(164, 860)$    & $87.3$ & $0.085$\\
    bottom-left object $(310, 80)$  & $(763, 164)$    & $(847, 164)$    & $84.2$ & $0.082$\\
    bottom-right object $(310, 420)$& $(763, 860)$    & $(847, 860)$    & $84.2$ & $0.082$\\
    \bottomrule
  \end{tabular}}
  \caption{Centroid displacement on a single $500\!\times\!375$
  PASCAL image. The image centre is invariant by symmetry. Every
  off-centre object centroid shifts by $\sim$8\,\% of the canvas
  along the padded axis. With SAM3-PCS's instance-localised
  prediction and the loader's letterboxed ground truth, this is
  enough to drop IoU to near-zero on small objects whose IoU support
  is narrow in the displaced direction.}
  \label{tab:alignment_example}
\end{table}

The 8\,\% number is not specific to this example: it is set by the
aspect ratio. For PASCAL VOC, where most images have aspect ratio
$\in[1.3, 1.5]$, off-centre objects shift by $5$--$10\,\%$ of the
canvas; for COCO and ADE-20K, which include both portrait and
landscape frames, the range is similar. The diagnostic returns the
same order of magnitude on all three. The HM gain reported in
Table~\ref{tab:alignment} ($+37.1$\,pp on PASCAL split-0, $-16.4$\,pp
under SAM2+CLIP) is the integrated effect of this displacement over
the full val set.

\section{ADE-OW Construction}
\label{sec:appendix_ade_ow_construction}

This appendix specifies the construction of ADE-OW, the held-out
26-class subset of ADE-20K used in Table~\ref{tab:alpha_ade}. The
artifact is shipped in machine-readable form alongside the code
release (\texttt{owgfss/benchmarks/ade\_ow/}); we summarise the
construction here for reproducibility.

\subsection{Filter rule}
\label{sec:appendix_ade_ow_filter}

ADE-OW retains every ADE150 class whose names share \emph{no}
WordNet noun-synset lemma with any label in the union of
COCO-stuff (181 thing+stuff labels), Pascal VOC (20 foreground
classes), and ImageNet-1k (1000 classes). Concretely:

\begin{enumerate}
  \item For each candidate ADE150 class, split its name on ``\verb|;|''
        into synonyms (e.g.\ \emph{streetlight;street lamp} $\to$
        \verb|{streetlight, street lamp}|).
  \item For each synonym, collect every WordNet noun-synset lemma
        reachable from any of its tokens (no hypernym/hyponym
        expansion).
  \item Form the union of these lemma sets over all synonyms of the
        candidate; call it $L_c$.
  \item Build the same lemma sets for every COCO-stuff, VOC, and
        IN-1k label, with one preprocessing exception: COCO-stuff
        compound labels of the form \texttt{\{class\}-other},
        \texttt{\{class\}-stuff}, \texttt{\{class\}-merged} are
        decomposed into their whitespace tokens (e.g.\
        \texttt{sky-other} contributes the bare lemma \emph{sky}).
        VOC and IN-1k labels are kept whole, so
        \texttt{table\_lamp} does \emph{not} leak the bare class
        \emph{lamp}.
  \item Drop the candidate iff $L_c$ intersects any of these three
        union lemma sets.
\end{enumerate}

The full filter is implemented in
\texttt{owgfss/benchmarks/ade\_ow/build\_ade\_ow.py} and runs in
seconds on CPU; the script writes the kept and rejected lists to
\texttt{ade\_ow\_classes.json} and \texttt{ade\_ow\_rejected.json}
respectively.

\subsection{Retained classes}
\label{sec:appendix_ade_ow_classes}

\Cref{tab:ade_ow_classes} lists the 26 surviving ADE150 classes with
their ADE-20K class indices, the synonyms used as text concepts at
inference, and the original ADE-20K val/train image counts (before
the $K\!=\!5$ support sampler removes some val images). Display
names for rows 18, 50, 74, 88, 97, 106, 113, 121, and 148 patch a
known upstream mangling of the
CSAILVision/sceneparsing CSV: rows whose original \texttt{Name}
field contained both ``\verb|;|'' (synonym separator) and a space
had spaces silently turned into semicolons. The filter logic is
robust to this because WordNet lookup operates on each whitespace
token, but the display name is patched for readability.

\begin{table}[t]
  \centering
  \footnotesize
  \setlength{\tabcolsep}{4pt}
  \begin{tabular}{@{}rlrrr@{}}
    \toprule
    \# & Class (synonyms separated by ``;'') & ADE idx & val n & train n\\
    \midrule
     1 & lamp                                     &  37 & 302 & 3089\\
     2 & streetlight;street lamp                  &  88 & 239 & 1989\\
     3 & box                                      &  42 & 162 & 1440\\
     4 & cushion                                  &  40 & 153 & 1453\\
     5 & sconce                                   & 135 & 108 & 1020\\
     6 & armchair                                 &  31 &  98 & 1172\\
     7 & column;pillar                            &  43 &  77 &  800\\
     8 & basket;handbasket                        & 113 &  75 &  622\\
     9 & awning;sunshade;sunblind                 &  87 &  61 &  533\\
    10 & chandelier;pendant;pendent               &  86 &  56 &  583\\
    11 & flag                                     & 150 &  56 &  421\\
    12 & stairway;staircase                       &  60 &  52 &  564\\
    13 & fan                                      & 140 &  44 &  397\\
    14 & fireplace;hearth;open fireplace          &  50 &  38 &  468\\
    15 & plaything;toy                            & 109 &  38 &  340\\
    16 & countertop                               &  71 &  31 &  331\\
    17 & canopy                                   & 107 &  31 &  292\\
    18 & sculpture                                & 133 &  21 &  285\\
    19 & tower                                    &  85 &  18 &  147\\
    20 & shower                                   & 146 &  14 &  130\\
    21 & kitchen island                           &  74 &   9 &  144\\
    22 & hovel;hut;hutch;shack;shanty             &  80 &   8 &   65\\
    23 & escalator;moving staircase;moving stairway &  97 &   6 &   42\\
    24 & lake                                     & 129 &   5 &   52\\
    25 & ship                                     & 104 &   4 &   52\\
    26 & conveyer belt;conveyor belt;transporter  & 106 &   4 &   57\\
    \bottomrule
  \end{tabular}
  \caption{The 26 ADE-OW classes, sorted by ADE-20K val frequency.
  ``val n'' / ``train n'' are ADE-20K val/train image counts before
  the $K\!=\!5$ support sampler. The three lowest-frequency rows
  ($n\!\le\!5$) have all of their val images consumed by the
  sampler at $K\!=\!5$ and therefore appear as $n\!=\!0$ in
  Table~\ref{tab:alpha_perclass}.}
  \label{tab:ade_ow_classes}
\end{table}

\subsection{Why 26 and not 30}
\label{sec:appendix_ade_ow_count}

An earlier draft of the protocol specification anticipated 30
surviving classes. The strict filter of
\cref{sec:appendix_ade_ow_filter} returns 26: relaxing the rule to
hit a round number would have introduced classes with known
PASCAL/COCO/ImageNet lemma overlap (the dominant family is
furniture nouns whose ADE compound names share a head lemma with a
plain ImageNet class, e.g.\ \emph{table\_lamp}/\emph{lamp} would
have leaked under a hypernym-permissive rule). We ship the 26
strict survivors as the canonical ADE-OW class list.

\subsection{Sampling protocol}
\label{sec:appendix_ade_ow_sampling}

Of ADE-20K's $983$-image val set, the val-image filter
(\texttt{build\_ade\_ow\_val\_filter.py}) keeps every val frame
containing $\ge\!1$ pixel of any ADE-OW class. The $K$-shot support
sampler picks the $K\!=\!5$ val images of highest per-class pixel
coverage as the support set for each ADE-OW class; the remaining
val images form the per-class query pool, with $N$ counted in
Table~\ref{tab:alpha_perclass}. The CLIP centroid is built once
per class from the $K\!=\!5$ support images using the same masked-foreground
operator $\phi$ as Eq.~\eqref{eq:centroid}, with no augmentation.
Three classes (\emph{lake},
\emph{ship}, \emph{conveyer belt}) have all of their val instances
consumed by the support sampler at $K\!=\!5$ and therefore appear
in Table~\ref{tab:alpha_perclass} with $n\!=\!0$ query images;
their rows are retained for vocabulary completeness.

\end{document}